\documentclass[letterpaper, 10 pt, conference]{ieeeconf}  % Comment this line out if you need a4paper
\IEEEoverridecommandlockouts
\usepackage{cite}
\usepackage{amsmath,amssymb,amsfonts}
\usepackage{algorithmic}
\usepackage{algorithm}
\usepackage{graphicx}
\usepackage{textcomp}
\usepackage{bbding}
\usepackage{booktabs}
\usepackage{multirow}
\usepackage{pifont}
\usepackage{tikz}
\usepackage{makecell}
\usepackage{setspace}
\usepackage{stfloats}
\usepackage{lipsum}
\usepackage{multirow}
\usepackage{bbding}
\usepackage[colorlinks,bookmarksopen,bookmarksnumbered,citecolor=green, linkcolor=red, urlcolor=black]{hyperref}
\usepackage{xcolor}
\def\BibTeX{{\rm B\kern-.05em{\sc i\kern-.025em b}\kern-.08em
    T\kern-.1667em\lower.7ex\hbox{E}\kern-.125emX}}

\title{\LARGE \bf
WildOcc: A Benchmark for Off-Road 3D Semantic Occupancy Prediction
}

\author{Heng Zhai$^{1,2,*}$, Jilin Mei$^{1,\dag}$, Chen Min$^{1}$, Liang Chen$^{1}$, Fangzhou Zhao$^{1}$, Yu Hu$^{1,3,\dag}$% <-this % stops a space
\thanks{\ddag This work was supported by National Natural Science Foundation of China under Grant No.U23B2034, No.62203424 and No.62176250, and the Innovation Program of Institute of Computing Technology, Chinese Academy of Sciences under Grant No. 2024000112.}% <-this % stops a space
\thanks{$^{1}$Research Center for Intelligent Computing Systems, Institute of Computing Technology, Chinese Academy of Sciences, Beijing, 100190, China.}%
\thanks{$^{2}$School of Information Science and Technology, ShanghaiTech University, Shanghai, 201210, China.}%
\thanks{$^{3}$School of Computer Science and Technology, University of Chinese Academy of Sciences, Beijing, 100190, China.}%
\thanks{$^{*}$Work done as an intern at Institute of Computing Technology, Chinese Academy of Sciences.}%
\thanks{$^{\dag}$Correspondence: Jilin Mei, Yu Hu, \{meijilin, huyu\}@ict.ac.cn}%
}

\begin{document}
\maketitle
\thispagestyle{empty}
\pagestyle{empty}

\begin{abstract}
3D semantic occupancy prediction is an essential part of autonomous driving, focusing on capturing the geometric details of scenes. Off-road environments are rich in geometric information, therefore it is suitable for 3D semantic occupancy prediction tasks to reconstruct such scenes. However, most of researches concentrate on on-road environments, and few methods are designed for off-road 3D semantic occupancy prediction due to the lack of relevant datasets and benchmarks. In response to this gap, we introduce WildOcc, to our knowledge, the first benchmark to provide dense occupancy annotations for off-road 3D semantic occupancy prediction tasks. A ground truth generation pipeline is proposed in this paper, which employs a coarse-to-fine reconstruction to achieve a more realistic result. Moreover, we introduce a multi-modal 3D semantic occupancy prediction framework, which fuses spatio-temporal information from multi-frame images and point clouds at voxel level. In addition, a cross-modality distillation function is introduced, which transfers geometric knowledge from point clouds to image features. Dataset will be released at \href{https://github.com/LedKashmir/WildOcc}{https://github.com/LedKashmir/WildOcc}
\end{abstract}

\section{Introduction}
The target of semantic occupancy prediction task is to predict the semantic labels of each occupied voxel at a distance, and making accurate and reliable 3D semantic occupancy prediction is challenging\cite{Li2023FBOCC3O,Wang2023OpenOccupancyAL,wei2023surroundocc}. As the ultimate target of autonomous driving, liberating driver implies that the car could autonomously drive in any road environment. Therefore, besides in on-road environments, it is necessary to study autonomous driving in off-road environments\cite{Min2022ORFDAD}. Off-road environments are rich in irregular objects, such as grass, bushes and trees\cite{Jiang2020RELLIS3DDD}. Therefore, it is a challenge to reconstruct 3D scene in complex off-road environments. Because the 3D semantic occupancy representation could provide more comprehensive information than 2D representation\cite{Wang2023OpenOccupancyAL}, studying 3D semantic occupancy prediction in off-road environments is of great importance.

\begin{figure}[t]
\centerline{\includegraphics[width=\linewidth]{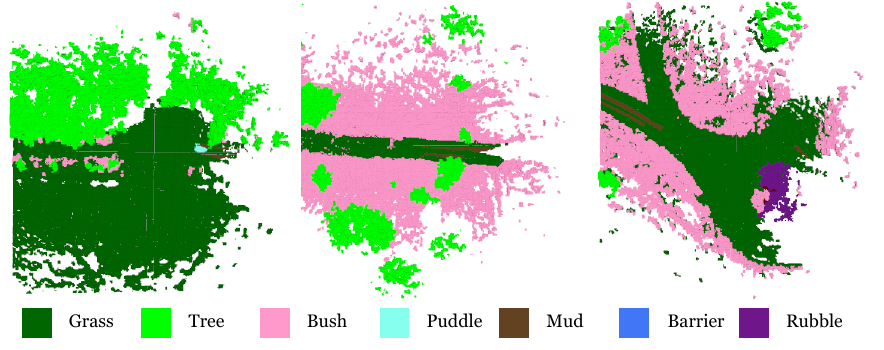}}
\caption{WildOcc provides dense semantic occupancy annotations for 10000 frames off-road dataset. To demonstrate the effect, here providing examples of large-scale annotations from top-view. (\textbf{Better viewed when zoomed in.})}
\label{fig:dataset}
\end{figure}

At present, there are growing researches on 3D semantic occupancy perception in on-road environments, such as TPVFormer\cite{Huang2023TriPerspectiveVF}, SurroundOcc\cite{wei2023surroundocc} and OpenOccupancy\cite{Wang2023OpenOccupancyAL}. However, to authors' knowledge, there are no researches on 3D semantic occupancy perception in off-road environments. Therefore, it is necessary to propose a dataset and benchmark of 3D semantic occupancy perception in off-road environments. \textbf{WildOcc}, the first benchmark of 3D semantic occupancy prediction in off-road environments, is proposed in this paper. We extends the off-road dataset Rellis-3D\cite{Jiang2020RELLIS3DDD} with dense 3D semantic occupancy labels, as shown in Fig.~\ref{fig:dataset} and Table~\ref{tab:dataset}. And we optimize the pipeline of annotations generation\cite{wei2023surroundocc} to fit off-road environments. Due to the amount of environmental objects such as trees and bushes, directly reconstructing the whole scene causes errors, as shown in Fig.~\ref{fig:match}. Therefore, a coarse-to-fine reconstruction is proposed to reconstruct the mesh of scenes more accurately (see Sec.\ref{sec:mesh}).

In addition, we introduce a 3D semantic occupancy prediction framework \textbf{OFFOcc} for off-road environments. OFFOcc transforms image features and LiDAR features into the unified voxel space, which reduces the loss of height information\cite{Wang2023OpenOccupancyAL} and provides feasibility for cross-modality distillation\cite{Li2022UnifyingVR}. In the camera branch and LiDAR branch, OFFOcc aligns and fuses features of historical frames at voxel level. Moreover, a cross-modality distillation function is proposed to extract geometric knowledge from LiDAR branch (Teacher) to camera branch (Student). Experiments have shown our method achieves high performance on the off-road 3D semantic occupancy prediction task. In short, the contributions of our work are summarized in following points:
\begin{itemize}
\item Our work applies 3D semantic occupancy prediction to off-road environments, and introduces WildOcc, to authors' knowledge, the first 3D semantic occupancy prediction benchmark in off-road environments.
\item A dense ground truth generation pipeline is introduced to reconstruct off-road environments more accurately, with a coarse-to-fine reconstruction.
\item We introduce a framework OFFOcc, with spatio-temporal alignment on  LiDAR and camera branch separately, and propose a novel distillation function to employ cross-modality distillation.
\end{itemize}

\begin{table*}[t!]
    \caption{\textbf{Comparison of WildOcc and other 3D semantic occupancy datasets}}
    \centering
    \resizebox{0.75\textwidth}{!}{%
    \begin{tabular}{@{}c|*{6}{c}@{}}
    \toprule
    \rule{0pt}{2ex}\multirow{1}{*}{Dataset} & \multirow{1}{*}{Type} & \multirow{1}{*}{Modality} & \multirow{1}{*}{Volume Size}& \multirow{1}{*}{\# Scenes} & \multirow{1}{*}{\#Frames}& \multirow{1}{*}{Resolution(m)}\\
    \midrule
    \rule{0pt}{2ex}SemanticKITTI\cite{Behley2019SemanticKITTIAD}& On-Road &C+L &[256, 256, 32]&22 &9K&[0.2, 0.2, 0.2]\\
    \rule{0pt}{2ex}Occ3D-Waymo\cite{Tian2023Occ3DAL}& On-Road & C+L&[200, 200, 16]&1K &200K&[0.4, 0.4, 0.4]\\
    \rule{0pt}{2ex}Occ3D-nuScenes\cite{Tian2023Occ3DAL}& On-Road & C+L&[200, 200, 16]&1K &40K&[0.4, 0.4, 0.4]\\
    \rule{0pt}{2ex}nuScenes-Occupancy\cite{Wang2023OpenOccupancyAL}& On-Road &  C+L&[512, 512, 40]&850 &34K&[0.2, 0.2, 0.2]\\
    \rule{0pt}{2ex}nuScenes-SurroundOcc\cite{wei2023surroundocc}& On-Road &  C+L&[200, 200, 16]&850 &34K&[0.5, 0.5, 0.5]\\
    \midrule
    \rule{0pt}{2ex}WildOcc(Ours)& Off-Road & C+L&[100, 100, 40]&5 &10K&[0.2, 0.2, 0.2]\\
    \bottomrule
\multicolumn{7}{l}{\textit{C, L} means camera and LiDAR.}
\end{tabular}}
\label{tab:dataset}
\end{table*}

\section{Related Work}

\subsection{3D Semantic Occupancy Prediction}
Recently, 3D semantic occupancy prediction is brought into focus. Several benchmarks\cite{Behley2019SemanticKITTIAD,Wang2023OpenOccupancyAL,wei2023surroundocc,Tian2023Occ3DAL} are released. In the field of autonomous driving, among camera-based methods, MonoScene\cite{Cao2021MonoSceneM3} is firstly proposed to predict semantic occupancy from a single camera. SurroundOcc\cite{wei2023surroundocc} extracts multi-scale image features from multi-cameras. PanoOcc\cite{Wang2023PanoOccUO} utilizes voxel queries to align spatial and temporal information. LiDAR-based method, such as UDNet\cite{fang2023udnet}, extracts voxel features by 3D U-Nets\cite{iek20163DUL}. Among multi-modal methods, OpenOccupancy\cite{Wang2023OpenOccupancyAL} proposes a benchmark derived from nuScenes and proposes several baselines. Co-Occ\cite{Pan2024CoOccCE} fuses features from LiDAR and camera with volume rendering regularization\cite{mildenhall2021nerf}. However, methods above are all in on-road environments. To authors’ knowledge, there are no researches on 3D semantic occupancy perception in off-road environments. Because the 3D semantic occupancy representation provides more comprehensive information than 2D representation\cite{zhang2024fusionocc}, we propose a benchmark WildOcc and a framework OFFOcc for 3D semantic occupancy prediction in off-road environments. 

\begin{figure*}[h!]
\centerline{\includegraphics[width=0.9\textwidth]{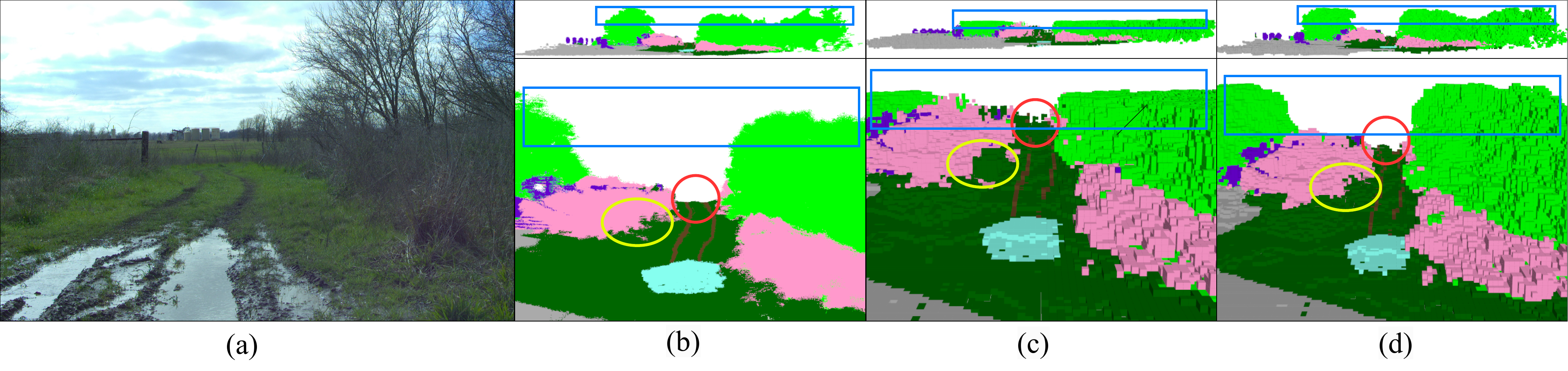}}
\caption{An example of results from different pipelines. (a) Image view of the scene. (b) Point clouds after multi-frame aggregation. (c) Annotations from pipeline of SurroundOcc\cite{wei2023surroundocc}, without coarse-to-fine reconstruction. (d) Annotations from pipeline of WildOcc (Ours), with coarse-to-fine reconstruction. Regions high-lighted by blue, yellow and red indicate that the coarse-to-fine reconstruction can make the annotations of off-road environments closer to real scene.}
\label{fig:match}
\end{figure*}

\subsection{Knowledge Distillation}
The initial target of knowledge distillation\cite{hinton2015distilling} is to compress model, by the way of transferring knowledge from teacher model to student model. Due to its effectiveness, knowledge distillation has been applied to semantic occupancy prediction\cite{Xia2023SCPNetSS,Zheng2024MonoOccDI}. SCPNet\cite{Xia2023SCPNetSS} transfers semantic knowledge from a multi-frame point clouds network to a single-frame point clouds network, the teacher and student employ the same architecture. MonoOcc\cite{Zheng2024MonoOccDI} utilizes knowledge distillation to transfer information from the teacher, the teacher model adopt a pre-trained larger backbone to extract rich features from multi-frame images. However, methods above transfer knowledge in the same modality. To utilize rich geometric information from LiDAR, our method proposes a cross-modality knowledge distillation strategy to transfer knowledge from LiDAR branch to camera branch.

\subsection{Off-road Perception}
Datasets of autonomous driving can be divided into two categories: on-road and off-road. On-road datasets are extensively used for autonomous driving research\cite{Behley2019SemanticKITTIAD,Caesar2019nuScenesAM,Sun2019ScalabilityIP}. However, off-road datasets are relatively scarce. DeepScene\cite{Yee2021DeepSceneSC} is the first public off-road dataset, RELLIS-3D\cite{Jiang2020RELLIS3DDD} utilizes multi-modal sensors to improve autonomous navigation in off-road conditions. ORFD\cite{Min2022ORFDAD} emphasizes traversability analysis, a crucial aspect for detecting freespace in off-road environments. Current off-road perception tasks mainly focus on freespace detection\cite{Min2022ORFDAD} and segmentation\cite{Jiang2020RELLIS3DDD}, there are few researches on reconstructing off-road environments. Because reconstructing scenes can provide more comprehensive perceptual information, it is important to employ 3D semantic occupancy task into off-road environments. Therefore, we design a benchmark WildOcc with accurate dense occupancy ground truth for off-road environments. Furthermore, a framework OFFOcc is introduced for off-road 3D semantic occupancy prediction.

% \section{WildOcc Benchmark}

\section{Pipeline of WildOcc}
\subsection{Multi-frame Point Clouds Aggregation}
Following SurroundOcc\cite{wei2023surroundocc}, given $P_{ego}^i$ as point clouds segment in ego coordinates of $i$-frame. We compute the world coordinates of $i$-frame as $P_w^i=T_w^i\cdot P_{ego}^i$, where $T_w^i$ is transform matrix consists of calibrated and ego-poses information. Then we concatenate multi-frame point clouds segments in world coordinates as $P_w=concat[P_w^1, ... , P_w^n]$, $n$ represent the number of frames in the sequence. Finally, the points of current frame are calculated, as $P=T_I\cdot P_w$, where $T_I$ represents transformation from world coordinates to current coordinates.
\subsection{Coarse-to-fine Reconstruction}
\label{sec:mesh}
Due to the interspace and uneven distribution of dense point clouds $P$, it is necessary to use Poisson Reconstruction\cite{Kazhdan2006PoissonSR} converting $P$ to a mesh. However, in our experiments, the reconstruction strategy of SurroundOcc\cite{wei2023surroundocc}, which reconstructs the whole scene directly, causes relatively big errors in off-road environments, as shown in Fig.~\ref{fig:match}. To fit the real scene well, we design a coarse-to-fine reconstruction. According to the semantic category of each point, we first divide points $P$ into ground points $P_g$ and non-ground points $P_{ng}$. Because ground elements have a relatively simple geometric structure, we use a coarse Poisson Reconstruction to reconstruct $P_g$ for ground mesh $\mathcal{M}_g$, the depth parameter of coarse Poisson Reconstruction is smaller. To the contrary, when facing non-ground elements, considering the complex geometric structure, we utilize a fine Poisson Reconstruction to get ground mesh $\mathcal{M}_{ng}$, with bigger depth parameter. Finally, we concatenate $\mathcal{M}_g$ and $\mathcal{M}_{ng}$ to get entire scene mesh $\mathcal{M}=concat[\mathcal{M}_g, \mathcal{M}_{ng}]$. The coarse-to-fine reconstruction can reconstruct the geometric structure more accurately, as shown in Fig.~\ref{fig:match}, and the coarse process is able to decrease the cost of computing. We voxelize the obtained mesh $\mathcal{M}$ in order to achieve dense voxels $V_{geo}$.

\subsection{Semantic Labeling}
After achieving dense voxels $V_{geo}$, the occupancy annotations still have no semantic information. The annotations of point clouds from Rellis-3D\cite{Jiang2020RELLIS3DDD} provides semantic information. Different from the method in SurroundOcc\cite{wei2023surroundocc}, which searches the nearest points for each voxel and assign the semantic label to it. We utilize KNN algorithm\cite{peterson2009k} to find the k-nearest points of each voxel center, and assign the most frequent category as semantic label of the voxel, which could reduce the interference of noise points. After this process, all voxels in $V_{geo}$ would get their semantic category. Compared with method in SurroundOcc\cite{wei2023surroundocc}, our method reduces the possibility of errors.

\begin{figure*}[h!]
\centerline{\includegraphics[width=0.9\textwidth]{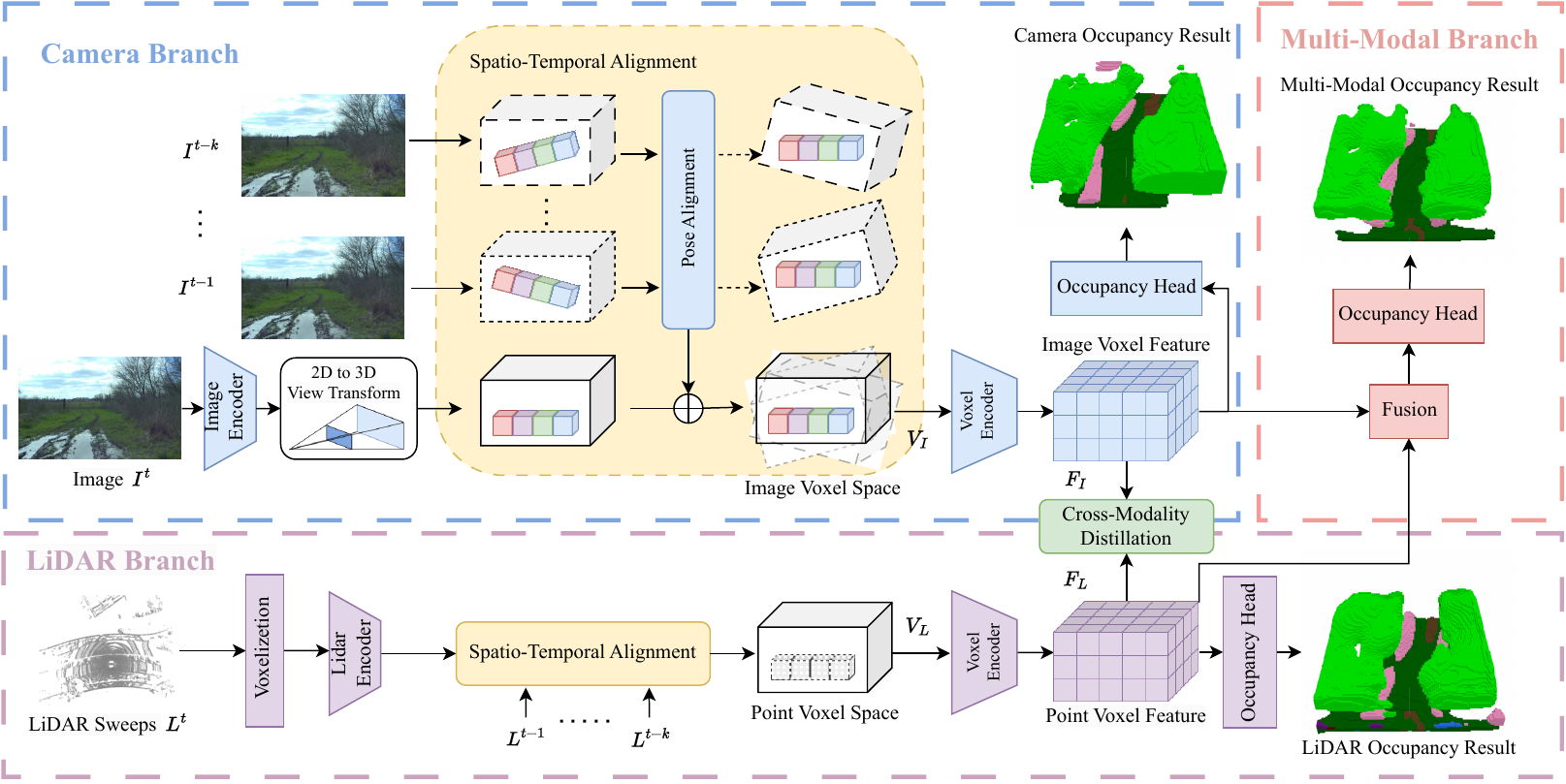}}
\caption{The overall architecture of framework OFFOcc. It consists of camera, LiDAR and multi-modal branches. To utilize the information of historical frames, we design a module of spatio-temporal alignment to combine the information. When training camera branch, we use LiDAR branch as the teacher and camera branch as the student, to transfer geometric knowledge from LiDAR branch.}
\label{fig:framework}
\end{figure*}

\section{OFFOCC METHOD}
The framework of OFFOcc, as shown in Fig.~\ref{fig:framework}. It consists of camera branch, LiDAR branch, and multi-modal branch. A spatio-temporal alignment module is proposed to utilize temporal information. Moreover, we introduce a cross-modality distillation function to transfer knowledge from LiDAR to camera. Specially, the camera branch and LiDAR branch can work independently, which ensures that the system still works when a single sensor is out-of-work. This design is necessary in off-road environments which are hard and unsafe.
\subsection{Task Definition}
The target of 3D semantic occupancy prediction is to generate the geometric and semantic representation of a 3D scene. Following the definition in 3D semantic occupancy prediction tasks\cite{Wang2023OpenOccupancyAL,Tian2023Occ3DAL}, we take $X^t$ and $X^{t-1},...,X^{t-k}$ as input, where $X^t=\{L^t, I^t\}$ represents current frame's LiDAR sweeps $L^t$ and image $I^t$, $X^{t-1},...,X^{t-k}$ denotes the data from previous frames, and $k$ represents the number of historical frames. The goal is to predict the occupancy labels $\mathcal{O}(X^t) \in \mathbb{R}^{H, W, D}$, where $H, W, D$ represents the volume size, and $\mathcal{O}(X^t)$ includes semantic classes and occupancy state (``empty" or ``not-empty") of each voxel.
\subsection{Camera Branch}
\subsubsection{Image Encoder \& 2D-to-3D Transformation}
The input of camera branch is a sequence of images $I^{t},...,I^{t-k}$. For current frame, we utilize ResNet\cite{He2015DeepRL} and FPN\cite{Lin2016FeaturePN} to extract the image features. Then, to achieve voxel representation of image, we adopt a 2D-to-3D transformation\cite{Wang2023OpenOccupancyAL} to project the 2D image features into 3D voxel space. The image voxel features of current frame is $V_I^t\in \mathbb{R}^{C,H,W,D}$, where $C$ is the dimensions of features and $H,W,D$ is the size of volume. The images of previous frames are processed as well, the total output of this module are $V_I^{t},...,V_I^{t-k}$.
\subsubsection{Spatio-temporal Alignment}
The effectiveness of temporal information has been proved in previous methods, such as \cite{Li2022BEVFormerLB, Park2022TimeWT, chen2024bevsoc}, which align features in BEV plane and cause the loss of height information. Due to the rugged surface of off-road environments, height changes frequently. Thus, we introduce a spatio-temporal alignment module, which utilize pose information to align features at voxel level. Utilizing the previous image voxel features $(V_I^{t-1}, ..., V_I^{t-k})$, for each voxel features $V_I^{t-i}$, we utilize the corresponding pose matrix $Pose^{t-i}$ from Rellis-3D\cite{Jiang2020RELLIS3DDD} to align features. The formulation of voxel alignment is:
\begin{equation}
V_I^{t-i\rightarrow w} = Pose^{t-i}\cdot V_I^{t-i},
\label{eq:pre2world}
\end{equation}
\begin{equation}
V_I^{w\rightarrow t} = (Pose^{t})^{-1}\cdot V_I^{t-i\rightarrow w}.
\label{eq:world2cur}
\end{equation}
Eq.(\ref{eq:pre2world}) represents transforming the voxel features of frame $t-i$ into world coordinates. Eq.(\ref{eq:world2cur}) denotes transforming voxel features from world coordinates into current frame's ego coordinates. For coordinate of each voxel in $V_I^{w\rightarrow t}$, we search the coordinate of nearest voxel in $V_I^{t}$ as new coordinate. After alignment, we concatenate voxel features of previous frames and current voxel features $V_I^{t}$ together as features of image voxel space $V_I$.
Then we utilize a voxel encoder to further extract features, and the output is image voxel features $F_I$:
\begin{equation}
F_I=VoxelEncoder(V_I),
\label{eq:voxel}
\end{equation}
where $F_I\in \mathbb{R}^{C,H,W,D}$, the voxel encoder consists of a 3D convolution layer, a BatchNorm layer and a ReLU layer.
\subsection{LiDAR Branch}
As illustrated in Fig.~\ref{fig:framework}, the LiDAR branch firstly utilizes voxelization\cite{Zhou2017VoxelNetEL} and LiDAR encoder\cite{Yan2018SECONDSE} to obtain the voxel features. Although the LiDAR sweep contains from multi-frame point clouds, it is aggregated without alignment. Because \textit{Spatio-Temporal Alignment} is designed for features at voxel level, it is suitable to use this module for aligning previous LiDAR voxel features as well, which is the same as alignment of camera branch. $V_L$ denotes the voxel features after alignment, where $V_L\in \mathbb{R}^{C,H,W,D}$. Then we use a voxel encoder as well, which outputs the point voxel features $F_L\in \mathbb{R}^{C,H,W,D}$. The size of point voxel features $F_L$ is same with image voxel features $F_I$. The unified representation of image voxel features and point voxel features makes cross-modality distillation possible.

\subsection{Multi-modal Branch}
The target of multi-modal branch is to improve the performance by utilizing all modalities, therefore we use the adaptive fusion module\cite{Wang2023OpenOccupancyAL} to dynamically fuse image voxel features $F_I$ and point voxel features $F_L$.
\begin{equation}
    F_M = \sigma(W) \odot F_L + \left( 1 - \sigma(W) \right) \odot F_I,
    \label{eq:f_formula}
\end{equation}
where $W$ is a trainable parameter matrix output by 3D convolution, $\odot$ means element-wise product and $\sigma$ is $\mathit{Sigmoid}$ function.

\subsection{Cross-modality Distillation}
Due to the complexity of off-road environments, sensors are at a risk of being damaged. One of our goals is to ensure the system still works when a single sensor is out-of-work, therefore it is necessary to enhance the ability of single modality. Because of the inherent characteristics of image and point clouds, point clouds help better utilize geometric structures in image. Due to the sparsity of point clouds, transferring detailed context from images is difficult. Therefore, we set the LiDAR branch as teacher network and the camera branch as student. To constrain the feature similarity only on occupied voxels, we propose a soft-supervision loss function as followed:
\begin{equation}
    \mathcal{L}_{\mathrm{distill}} = \frac{1}{H \times W \times D}
    \sum_{x}^{H} \sum_{y}^{W}\sum_{z}^{D}\mathcal{F}(x,y,z),
\end{equation}
\begin{equation}
        \mathcal{F}(x,y,z) = M_{x,y,z} \frac{\mathbf{f}_I(x,y,z)^\top \mathbf{f}_L(x,y,z)}{\|\mathbf{f}_I(x,y,z)\|_2 \|\mathbf{f}_L(x,y,z)\|_2},
\end{equation}
where $\mathbf{f}_I(x,y,z)\in \mathbb{R}^C$ is the features indexed as $(x,y,z)$ in the image voxel features $F_I$, $\mathbf{f}_L(x,y,z)$ is from LiDAR voxel features $F_L$ as well, and $M_{x,y,z} = 1$ if the annotation of $(x,y,z)$ is occupied (non-empty and non-noise), otherwise $M_{x,y,z} = 0$.

\subsection{Training Loss}
In the method above, we use multiple loss functions to supervise the network. For the knowledge transferring, we introduce distillation loss $\mathcal{L}_{\mathrm{distill}}$. For the occupancy prediction, we use cross-entropy loss $\mathcal{L}_{\text{ce}}$ and lovasz-softmax loss $\mathcal{L}_{\text{ls}}$\cite{Wang2023OpenOccupancyAL}. The total function as follows:
\begin{equation}
    \mathcal{L}_{\text{total}} = \mathcal{L}_{\text{ce}} + \mathcal{L}_{\text{ls}} + \lambda \mathcal{L}_{\mathrm{distill}} + \mathcal{L}_{\text{d}},
\end{equation}
where $\lambda$ is hyper-parameters, $\mathcal{L}_{\text{d}}$ is the depth loss\cite{Wang2023OpenOccupancyAL} which only works in camera or multi-modal branch. $\mathcal{L}_{\mathrm{distill}}$ only works when training the camera branch.

\definecolor{dirt}{rgb}{0.43, 0.08, 0.54}
\definecolor{grass}{rgb}{0, 0.5, 0}
\definecolor{tree}{rgb}{0, 1, 0} % 自定义绿色
\definecolor{bush}{rgb}{1, 0.6, 0.8} % 自定义绿色
\definecolor{Barrier}{rgb}{0.16, 0.5, 1} % 自定义绿色
\definecolor{Puddle}{rgb}{0.52, 1, 0.94} % 自定义绿色
\definecolor{Mud}{rgb}{0.39, 0.26, 0.13} % 自定义绿色

\begin{table*}[t!]
    \caption{\textbf{3D Semantic Occupancy Prediction Results on WildOcc test set.}}
    \centering
    \resizebox{0.9\textwidth}{!}{%
    \begin{tabular}{@{}c|c|cc|*{7}{c}|c|c@{}}
    \toprule
    \rule{0pt}{2ex}\multirow{1}{*}{Method} & \multirow{1}{*}{Modality} & \multirow{1}{*}{IoU} & \multirow{1}{*}{mIoU} & \multirow{1}{*}{\tikz\draw[fill=grass,draw=grass] (0,0) rectangle (0.2cm,0.2cm);Grass} & \multirow{1}{*}{\tikz\draw[fill=tree,draw=tree] (0,0) rectangle (0.2cm,0.2cm);Tree}& \multirow{1}{*}{\tikz\draw[fill=bush,draw=bush] (0,0) rectangle (0.2cm,0.2cm);Bush}&\multirow{1}{*}{\tikz\draw[fill=Puddle,draw=Puddle] (0,0) rectangle (0.2cm,0.2cm);Puddle}& \multirow{1}{*}{\tikz\draw[fill=Mud,draw=Mud] (0,0) rectangle (0.2cm,0.2cm);Mud}& \multirow{1}{*}{\tikz\draw[fill=Barrier,draw=Barrier] (0,0) rectangle (0.2cm,0.2cm);Barrier} & \multirow{1}{*}{\tikz\draw[fill=dirt,draw=dirt] (0,0) rectangle (0.2cm,0.2cm);Rubble}&
    \multirow{1}{*}{Input Size} & \multirow{1}{*}{2D Backbone}\\
    \midrule
    \rule{0pt}{2ex}SurroundOcc\cite{wei2023surroundocc}& C & 28.7 &10.2 &23.7&20.4&20.5&0.4&3.7&0.3&2.2&$900\times1600$&R101-DCN\\
    \rule{0pt}{2ex}OccFormer\cite{Zhang2023OccFormerDT}& C & 27.2 & 10.5&24.3 &21.1&20.7&0.5&2.3&0.7&\underline{3.6}&$896\times1600$&R101\\
    \rule{0pt}{2ex}C-CONet\cite{Wang2023OpenOccupancyAL}& C & 23.6 & 8.8&23.5 &10.8&19.2&0.3&\underline{5.8}&0.4&1.7&$896\times1600$&R101\\
    \rule{0pt}{2ex}FB-Occ\cite{Li2023FBOCC3O}& C & 28.4 & 10.6&24.1&21.6&20.4&\underline{1.1}&3.2&\underline{1.0}&2.5&$896\times1600$&R101\\
    \rule{0pt}{2ex}C-OFFOcc(Ours)& C & \underline{29.7} & \underline{11.2}&\underline{24.6} &\underline{23.8}&\underline{22.1}&0.6&3.5&0.6&3.2&$896\times1600$&R101\\
    \midrule
    \rule{0pt}{2ex}L-CONet\cite{Wang2023OpenOccupancyAL}& L & 32.4 & 11.0&25.3 &26.7&23.4&0.1&1.4&0.9&0.4&-&-\\
    \rule{0pt}{2ex}M-CONet\cite{Wang2023OpenOccupancyAL}& M & 31.6 &13.6 &27.2 &30.5&26.3&0.5&\textbf{7.1}&1.2&2.3&$896\times1600$&R101\\
    \rule{0pt}{2ex}L-OFFOcc(Ours)& L & 35.3 & 11.6 &22.8 &28.5&25.0&0.5&3.4&1.1&0.1&-&-\\
    \rule{0pt}{2ex}M-OFFOcc(Ours)& M & \textbf{32.8} & \textbf{14.8}&\textbf{28.6} &\textbf{33.4}&\textbf{27.5}&\textbf{0.9}&6.8&\textbf{1.7}&\textbf{4.6}&$896\times1600$&R101\\
    \bottomrule
\multicolumn{13}{l}{C-* denotes camera method, L-* denotes LiDAR method and M-* denotes camera-LiDAR fusion method.}\\
\multicolumn{13}{l}{C represents the modality is camera, L represents the modality is LiDAR and M represents the modality is camera-LiDAR fusion.} \\
\multicolumn{13}{l}{\underline{Underline} means the best performance of camera method, and \textbf{bold} means the best performance of camera-LiDAR fusion method.}
\end{tabular}}

\label{tab:comparison}
\end{table*}
\section{Experiment}
\subsection{Dataset}
Experiments are conducted on WildOcc dataset, which is proposed in this work. WildOcc is split into a training set with 7500 frames, a validation set with 1250 frames, and a test set with 1250 frames. The method is evaluated on the test set of WildOcc. Because the data collector is a diminutive mobile robot\cite{Jiang2020RELLIS3DDD} with monocular camera and LiDAR, we set the occupancy prediction range as $[0, 20m]$ for $X$ axis, $[-10m, 10m]$ for $Y$ axis and $[-2m, 6m]$ for $Z$ axis. The shape of occupancy is $100\times 100\times 40$ with voxel size of 0.2m. The test set has 7 semantics categories.

\subsection{Evaluation Metrics}
Following on-road occupancy benchmarks\cite{wei2023surroundocc,Wang2023OpenOccupancyAL,Tian2023Occ3DAL}, the setting of off-road evaluation metrics is the same with these benchmarks:
\begin{equation}
\text{IoU} = \frac{TP}{TP + FP + FN},
\end{equation}
\begin{equation}
\text{mIoU} = \frac{1}{N} \sum_{i=1}^{N} \frac{TP_i}{TP_i + FP_i + FN_i},
\end{equation}
where $TP, FP, FN$ represent the number of true positive, false positive and false negative voxel prediction. $N$ is the number of semantic classes.

\subsection{Implementation Details}
\subsubsection{WildOcc Generation}
WildOcc is derived from 5 scenes of Rellis-3D\cite{Jiang2020RELLIS3DDD}. In each scene, consecutive 2000 frames of point clouds are collected to form WildOcc with 10000 frames. For each scene, the first 1500 frames are set as training set, the middle 250 frames are set as validation set and the last 250 frames are set as test set. During multi-frame aggregation, we stitch consecutive 400 frames at once. The depth parameters of coarse and fine Poisson Reconstruction are set as 8 and 13 separately. $k$ is set to be 15 in the period of semantic labeling. 
\subsubsection{OFFOcc Implementation}
In the camera branch, raw images are resized into $896\times 1600$. ResNet101\cite{He2015DeepRL} and FPN\cite{Lin2016FeaturePN} are utilized as the image encoder. To generate dense depth maps, LiDAR points are projected onto the image, followed by performing depth completion\cite{Ku2018InDO} to make dense result. For LiDAR branch, the input are 10 LiDAR sweeps. We voxelize the sweeps and utilize a voxel encoder to extract features. The spatio-temporal alignment aggregates 4 frames (including current frame) with a 0.5s time interval. And the structure of occupancy head is the same in each branch. In training period, the LiDAR branch is trained firstly and frozen as teacher network after achieving the best performance. When training the camera branch, distillation function is utilized to transfer knowledge. And the hyper-parameters $\lambda$ is set as 0.8 in our experiments. Finally, the multi-modal branch is trained. Each branch of the model is trained for 20 epochs, with a batch size of 4 across 4 RTX 6000Ada GPUs. We utilize the AdamW\cite{Kingma2014AdamAM} optimizer with a weight decay of 0.01 and an initial learning rate of 3e-4.

\begin{figure*}[h!]
\centerline{\includegraphics[width=0.9\textwidth]{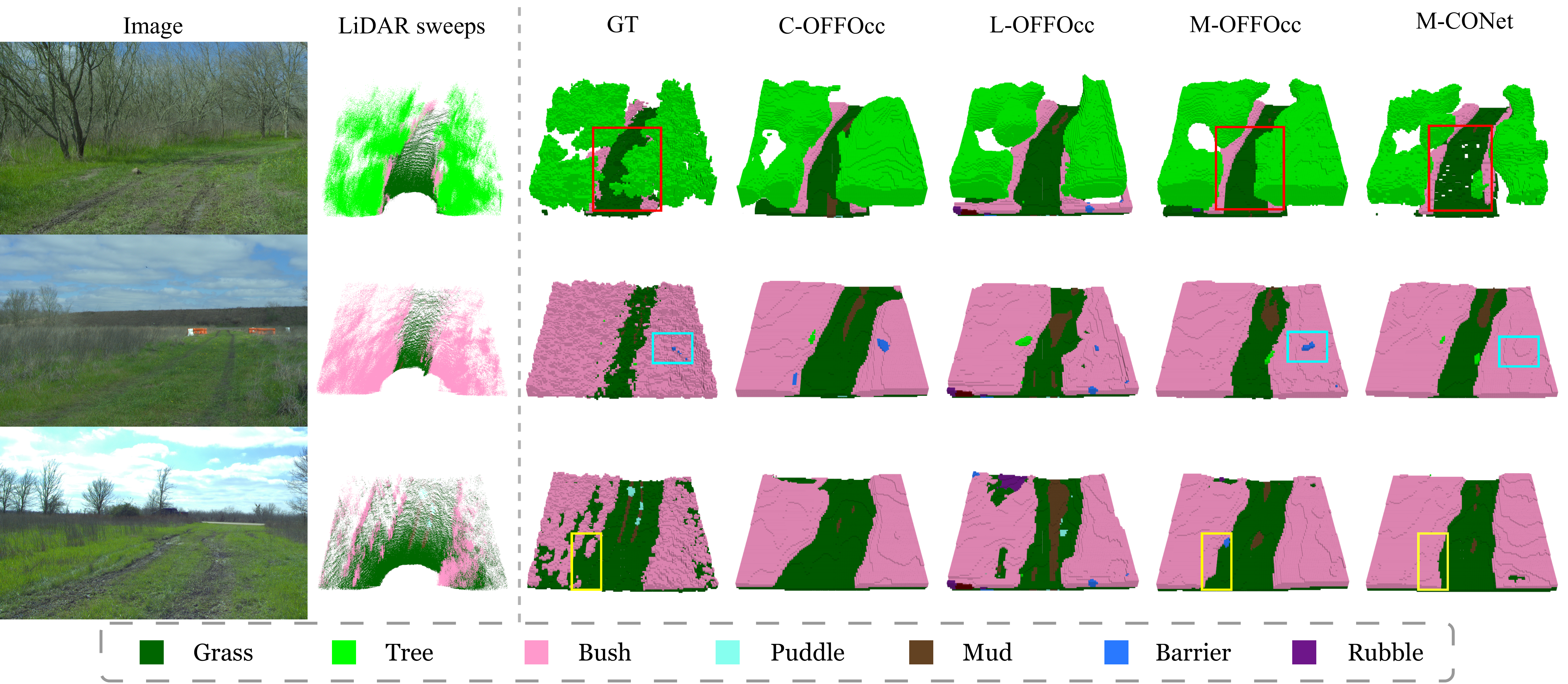}}
\caption{Qualitative results of OffOcc and M-CONet on WildOcc dataset. The input monocular image and LiDAR sweeps are shown on the left. (\textbf{Colors of LiDAR sweeps are convenient for demonstration, the actual LiDAR sweeps input does not contain semantic information. Better viewed when zoomed in.})}
\label{fig:result}
\end{figure*}
\subsection{Main Results}
Comparisons with several on-road methods on various modalities, as shown in Table~\ref{tab:comparison}. These methods use surround-view images as input, for fairly conducting experiments, we transform the input setting from surround-view into front-view. Further, the perception range and voxel size all follow the setting of WildOcc. Table~\ref{tab:comparison} demonstrates C-OFFOcc achieves the highest mIoU among camera-only methods. With geometric knowledge from LiDAR branch, C-OFFOcc has a increase of \textbf{$2.2\%$} IoU on the tree category compared to FB-Occ\cite{Li2023FBOCC3O} and \textbf{$1.4\%$} IoU on the bush category compared to OccFormer\cite{Zhang2023OccFormerDT}. These two categories contains rich geometric and height information. In addition, L-OFFOcc displays better performance than L-CONet\cite{Wang2023OpenOccupancyAL} on mIoU, and the mIoU of M-OFFOcc is larger than M-CONet\cite{Wang2023OpenOccupancyAL} as well. In addition, as shown in Fig.~\ref{fig:result}, the qualitative results demonstrate that M-OFFOcc predicts a more accurate 3D semantic occupancy prediction in off-road environments than M-CONet\cite{Wang2023OpenOccupancyAL}, and the results of M-CONet may contain interspace. Moreover, the results of M-OFFOcc are closer to the ground truth.

\begin{table}[t!]
\centering
\caption{Ablation study on the effectiveness of proposed component}
\resizebox{0.85\linewidth}{!}{%
\begin{tabular}{@{}*{4}{c}|cc@{}}
\toprule
\rule{0pt}{2ex}
\multirow{1}{*}{row} & Modality& \makecell{Temporal\\Alignment} &  \makecell{Cross-Modality\\Distillation} &  Train Mem. &mIoU↑ \\ 
\midrule
1 & C& \XSolidBrush &  \XSolidBrush  & 17G & 9.2\\ 
2 & C& \XSolidBrush &  \Checkmark  & 17G &  10.4\\ 
3 & C& \Checkmark &  \XSolidBrush  & 20G &  10.7\\ 
4 & C& \Checkmark &  \Checkmark  & 20G & \textbf{11.2} \\ 
\midrule
5 & L& \XSolidBrush &  -  & 6G &  10.2\\ 
6 & L& \Checkmark &  -  & 8G &  \textbf{11.6}\\ 
\midrule
7 & C+L& \XSolidBrush &  -  & 19G &  12.9\\ 
8 & C+L& \Checkmark &  -  & 23G &  \textbf{14.8}\\ 
\bottomrule
\multicolumn{6}{l}{Train Mem. represents the usage of GPU memory.} \\
\end{tabular}}
\label{tab:ablation}
\end{table}

\subsection{Ablation Study}
To verify the effectiveness of temporal alignment and components combination, we conduct several ablation studies, as illustrated in Table~\ref{tab:ablation}. The Temporal Alignment column represents whether to use spatio-temporal alignment. In our experiment setting, we only set camera branch as the student network. The Cross-Modality Distillation column represents whether to employ knowledge distillation. Furthermore, we verify the effectiveness of our soft-supervision distillation function, as shown in Table~\ref{tab:abdistill}. For fair experiments, when comparing distillation functions, we do not employ temporal alignment. 
\subsubsection{Spatio-temporal Alignment}
In Table~\ref{tab:ablation}, the comparison between row 1 and row 3 shows the effectiveness of spatio-temporal alignment in camera branch. It enhances the performance on WildOcc dataset by 1.2 mIoU. The differences between row 5 and row 6, row 7 and row 8 demonstrate the effectiveness in LiDAR and multi-modal branch as well.
\subsubsection{Cross-modality Distillation}
By comparing row 1 and row 2, row 3 and row 4 in Table~\ref{tab:ablation}, the performance increases 1.2 mIoU and 0.5 mIoU respectively. The improvement represents that the strategy of cross-modality is valid. In Table~\ref{tab:abdistill}, the results between different distillation functions demonstrate our soft-supervision distillation function improve the performance of cross-modality distillation as well.
\begin{table}[t!]
\centering
\caption{Ablation study on different distillation function}
\resizebox{0.7\linewidth}{!}{%
\begin{tabular}{@{}*{3}{c}|c@{}}
\toprule
\rule{0pt}{2ex}
\multirow{1}{*}{row} & Modality &  \makecell{Distillation Function} &mIoU↑ \\ 
\midrule
1 & C&  - & 9.2\\ 
2 & C&  $\mathcal{L}_2$ & 9.6\\ 
3 & C&   Partial $\mathcal{L}_2$\cite{Heo2019ACO} &  10.1\\ 
4 & C&   $KL$\cite{Cao2021MonoSceneM3} &  9.9\\ 
5 & C&   $\mathcal{L}_{\mathrm{distill}}$(Ours) &  \textbf{10.4}\\ 
\bottomrule
\end{tabular}}
\label{tab:abdistill}
\end{table}
\section{Conclusion}
In this paper, we propose the off-road 3D semantic occupancy benchmark WildOcc. A new pipeline is designed to generate dense occupancy annotations, and the dense annotations is closer to real scenes. In addition, we introduce a method OFFOcc to predict occupancy. The camera branch and LiDAR branch can work independently. For the camera branch and LiDAR branch, we design a spatio-temporal alignment module to utilize temporal information from historical frames at voxel level. A cross-modality distillation function is proposed to transfer geometric information from LiDAR to camera. Owing to these improvements, our model OFFOcc achieves high performance on off-road benchmark WildOcc. There still exists some limitations in our work. The small occupancy prediction range (20m) is one of current limitation, mainly due to the use of a diminutive mobile robot. The relatively small frame count is another. In the future work, we plan to construct an off-road dataset with equal distribution and large size, collected by a vehicle.

\newpage

\bibliographystyle{IEEEtran}
\bibliography{IEEEabrv,IEEEexample}

\end{document}